%% file: main.tex
\documentclass[10pt,twocolumn,letterpaper]{article}

\PassOptionsToPackage{table,dvipsnames}{xcolor}
\usepackage[pagenumbers]{cvpr} 

\usepackage[accsupp]{axessibility} 
\usepackage{adjustbox}
\usepackage{graphicx}
\usepackage{subcaption}
\usepackage{multirow}
\usepackage{svg}
\usepackage{hhline}
\usepackage{amsmath}   
\usepackage{amssymb}   
\usepackage{mathrsfs}
\usepackage{bm} 
\usepackage{makecell} 
\usepackage{colortbl}

\definecolor{dg}{rgb}{0,0.694,0.298}
\definecolor{purple}{rgb}{0.4,0.176,0.569}
\definecolor{royalblue}{RGB}{65,105,225}
\usepackage{pifont}
\newcommand{\figref}[1]{Fig.~\ref{#1}}

\newcommand{\secref}[1]{Sec.~\ref{#1}}
\newcommand{\tableref}[1]{Table~\ref{#1}}

\definecolor{lightgray}{RGB}{245,245,245}
\definecolor{cvprblue}{rgb}{0.21,0.49,0.74}
\usepackage[colorlinks,allcolors=cvprblue]{hyperref}
\def\paperID{10536} 
\def\confName{CVPR}
\def\confYear{2025}

\title{DPSeg: Dual-Prompt Cost Volume Learning for Open-Vocabulary Semantic Segmentation
}

\author{
    Ziyu Zhao$^{1*}$, Xiaoguang Li$^1$\thanks{Co-first authors and contribute equally.}, Linjia Shi$^1$, Nasrin Imanpour$^1$, Song Wang$^2$\thanks{Corresponding author.} \\
    $^1$University of South Carolina, USA \quad
    $^2$Shenzhen University of Advanced Technology, China \\
    {\tt\small \{ziyuz, xl22, linjia, imanpour\}@email.sc.edu, wangsong@suat-sz.edu.cn}
}

\begin{document}
\maketitle
\input{sec/0_abstract}    
\input{sec/1_intro}
\input{sec/2_relatework}

\input{sec/3_Discussion}

\input{sec/4_Method}
\input{sec/5_Experiment}

\input{sec/6_Conclusion}
{
    \small
    \bibliographystyle{ieeenat_fullname}
    \bibliography{main}
}


\end{document}

%% file: sec/0_abstract.tex
\begin{abstract}
Open-vocabulary semantic segmentation aims to segment images into distinct semantic regions for both seen and unseen categories at the pixel level. Current methods utilize text embeddings from pre-trained vision-language models like CLIP but struggle with the inherent domain gap between image and text embeddings, even after extensive alignment during training. Additionally, relying solely on deep text-aligned features limits shallow-level feature guidance, which is crucial for detecting small objects and fine details, ultimately reducing segmentation accuracy.
To address these limitations, we propose a dual prompting framework, DPSeg, for this task. Our approach combines dual-prompt cost volume generation, a cost volume-guided decoder, and a semantic-guided prompt refinement strategy that leverages our dual prompting scheme to mitigate alignment issues in visual prompt generation. By incorporating visual embeddings from a visual prompt encoder, our approach reduces the domain gap between text and image embeddings while providing multi-level guidance through shallow features. Extensive experiments demonstrate that our method significantly outperforms existing state-of-the-art approaches on multiple public datasets.
\end{abstract}

%% file: sec/1_intro.tex
\section{Introduction}
\label{sec:intro}

Semantic segmentation, the process of assigning class labels to individual pixels in an image, has seen significant advancements. Conventional segmentation methods \cite{long2015fully,krizhevsky2012imagenet,simonyan2014very,lin2017feature, chen2017deeplab,chen2016attention,zhao2022crossmodal,zhao2024crossmodal} are often structured for closed-set tasks, wherein models are trained and evaluated on a fixed set of categories. While these models demonstrate strong performance in controlled environments, they typically struggle to generalize effectively to real-world scenarios that include unseen objects and novel categories.


\begin{figure}[t]
\centering
\includegraphics[width=1\linewidth]{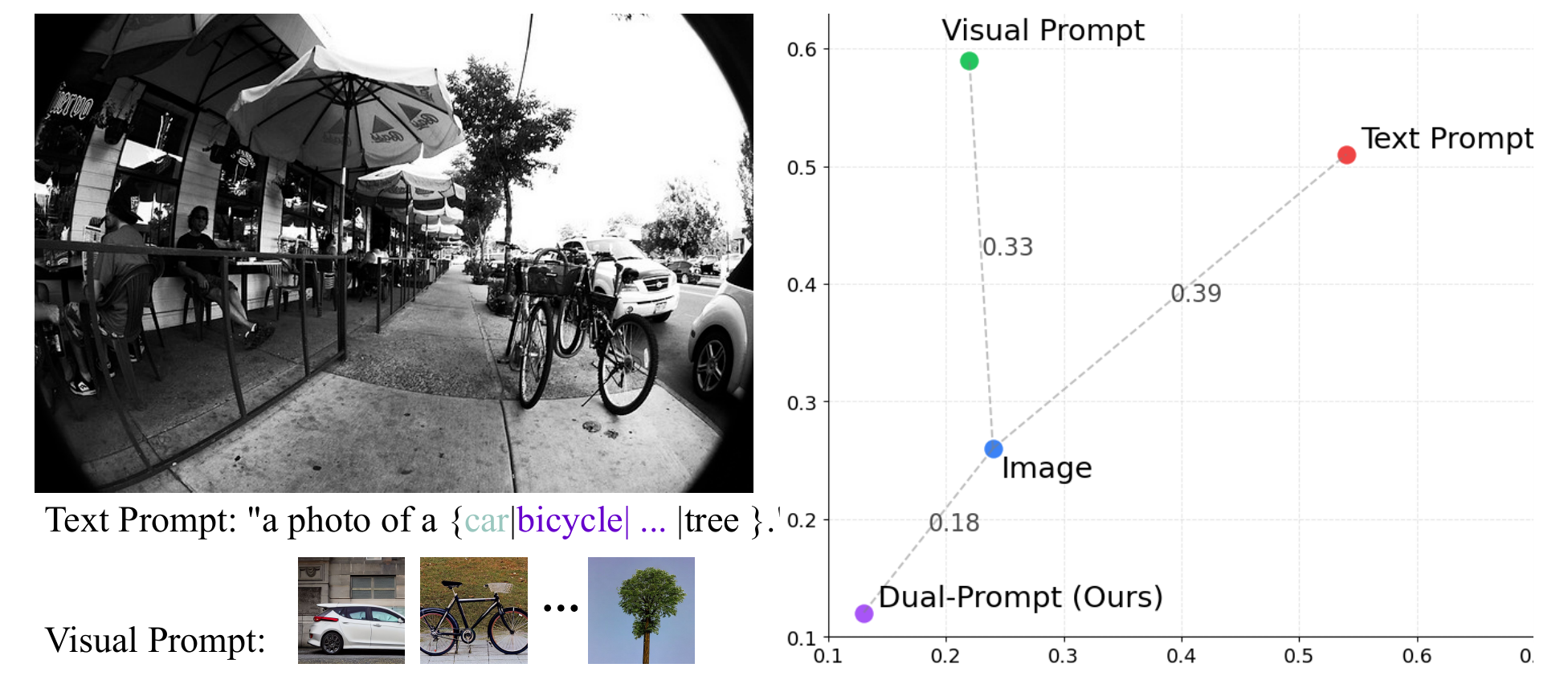}
\caption{\textbf{T-SNE visualization of embeddings in CLIP feature space, showing {text prompt}, {visual prompt}, {target image}, and our {dual-prompt} for a street scene image.} Numbers indicate distance to image feature, where our dual-prompt approach achieves closest proximity (0.18).}
\vspace{-10pt}
\label{tsne}
\end{figure}


Open-vocabulary semantic segmentation (OVSS) has emerged as a promising solution to this limitation. By leveraging large-scale vision-language models like CLIP~\cite{radford2021learning}, OVSS enables segmentation beyond fixed categories, allowing models to generalize to arbitrary classes during inference. Recent approaches, such as CAT-Seg,~\cite{cho2024cat} utilize pixel-level cost volumes---dense similarity mappings between local image features and text embeddings in CLIP's space---to balance segmentation accuracy and efficiency. SED~\cite{xie2024sed} further enhances this framework by incorporating a hierarchical encoder and multi-scale features. However, existing methods face two key challenges: (1) despite incorporating multi-scale semantic information, the reliance on upsampling the cost volume leads to potential degradation of fine-grained details, particularly affecting small but crucial semantic segments, and (2) the sole dependence on text-image embedding alignment fails to fully resolve the domain gap, even with robust pre-trained models.

To overcome these limitations, our work begins by conducting comprehensive experiments (see \secref{sec:domain-gap} and \secref{sec:cost-volume}) to investigate the domain gap between image and text embeddings by calculating cosine similarity scores across sampled categories. Our findings reveal that visual prompts exhibit stronger feature alignment with image representations in the embedding space compared to text prompts. Additionally, the cost volumes generated from visual prompts demonstrate richer spatial-semantic cues for precise segment localization than their text-based counterparts. Building on these insights, we propose a novel OVSS framework with two key components: (1) a dual-prompt cost volume generation mechanism that leverages both visual and text prompts, and (2) a cost volume-guided decoder that progressively integrates visual prompt features at corresponding feature levels for refined segmentation prediction. To further improve segmentation accuracy and mitigate potential misalignment between visual prompts and actual segments during inference, we propose a semantic-guided prompt refinement strategy that employs a two-stage process, where initial segmentation results serve as refined visual prompts to guide the final prediction. As visualized in \figref{tsne}, our dual-prompt scheme achieves superior embedding space alignment with a distance of 0.18 to image features, compared to 0.33 for visual prompts and 0.39 for text prompts alone. The t-SNE visualization demonstrates that our dual-prompt approach effectively bridges the semantic gap by integrating synergistic information from both modalities. Extensive experiments show that our method sets a new benchmark in open-vocabulary semantic segmentation, achieving superior accuracy, inference efficiency, and overall performance.
Our main contribution is the following:
\begin{itemize}

    \item Through both quantitative and qualitative analysis of CLIP embeddings, we demonstrate that visual prompts achieve stronger feature alignment and generate more precise spatial-semantic cost volumes compared to text-based approaches.

    \item We propose \textbf{DPSeg}, a novel approach for open-vocabulary semantic segmentation featuring dual-prompt cost volume generation and cost volume-guided decoder. Additionally, our semantic-guided prompt refinement module leverages initial predictions as scene-specific visual prompts to enhance segmentation quality through a two-stage inference process.

    \item The extensive experiment results on multiple open vocabulary semantic segmentation datasets demonstrate that our method outperforms competitors by a large margin.
    
\end{itemize}


%% file: sec/2_relatework.tex
\section{Related Work}
\label{sec:formatting}

\subsection{Cross-Modality Learning Models}


Cross-modality representation learning has seen significant advancements, broadly categorized into Spatial Cross Modality (2D-3D) \cite{zhao2024crossmodal} and Conceptual Cross Modality (Vision-Language). Vision-language models \cite{chen2020uniter, lu2019vilbert, tan2019lxmert, schuhmann2022laion} have been developed and fine-tuned on various downstream tasks using image-text pairs. CLIP \cite{radford2021learning, yao2021filip}, for instance, leverages large-scale image-text data to align visual and language representations, excelling in zero-shot image recognition and expanding into tasks like segmentation \cite{he2023clip,liang2023open,xu2023open,xu2023side}, captioning \cite{mokady2021clipcap}, classification \cite{abdelfattah2023cdul}, restoration \cite{zhang2024sair}, and object detection \cite{gu2021open,zhong2022regionclip}. Recent CLIP-based extensions have shown notable improvements across various tasks. SLIP \cite{mu2022slip} enhances CLIP by incorporating self-supervised image-to-image contrastive learning alongside CLIP’s image-text alignment, yielding richer and more robust visual representations that capture subtle image features more effectively. A-CLIP \cite{yang2023attentive} introduces an attentive token removal strategy, retaining only a small subset of tokens that exhibit strong semantic correlations with the corresponding text descriptions, allowing for more focused and relevant visual-text alignment.

In segmentation, OVSeg \cite{liang2023open} adapts CLIP by fine-tuning on masked image regions to produce mask-aware embeddings for object-level segmentation. However, it struggles to capture detailed semantic information, such as object attributes and relationships, reducing its effectiveness in complex scenes. Our model addresses this by leveraging higher intra-modal similarity through integrated visual prompts, enriching cross-modal cosine similarity and supplementing semantic information beyond text prompts alone.

\subsection{Open-Vocabulary Semantic Segmentation}

Open-vocabulary semantic segmentation (OVSS) enables models to segment arbitrary object categories, including unseen classes. While early zero-shot approaches relied on attribute-based classifiers~\cite{lampert2013attribute} and word embeddings~\cite{zhang2016zero}, vision-language models like CLIP~\cite{radford2021learning} and ALIGN~\cite{jia2021scaling} have significantly advanced the field through image-text alignment in shared embedding spaces. OVSS methods generally follow either \ding{182} a two-stage pipeline~\cite{ding2022open,huynh2022open,liang2023open,xu2022simple,ding2022decoupling,ghiasi2022scaling} or \ding{183} an end-to-end network~\cite{xu2023open,xu2022groupvit,xu2023side,zou2023generalized}.
Among two-stage approaches, SCAN~\cite{liu2024open} calibrates CLIP with semantic priors, EBSeg~\cite{shan2024open} employs balanced decoders with semantic consistency, and USE~\cite{wang2024use} leverages multi-granularity segments. End-to-end methods have progressed from MaskCLIP's~\cite{zhou2022extract} direct CLIP adaptation to SAN's~\cite{xu2023side} sophisticated side-adapter network. Recent advances like CAT-Seg~\cite{cho2024cat} and SED~\cite{xie2024sed} utilize cost volumes with auxiliary backbones and hierarchical feature fusion, respectively.
Building upon these works, we address the limitation of detail degradation during cost volume upsampling by proposing a multi-scale approach that combines both visual and text prompts to better utilize feature information across scales. 
\subsection{Visual Prompting}

Prompting generally refers to providing textual input to an AI model to elicit a specific task. Visual prompting extends this concept by using an image as a prompt, where semantically related reference images guide the model in tasks such as segmentation \cite{zhao2024crossmodal}. Visual prompting can be broadly categorized into prototype-based and feature-matching approaches. Prototype-based methods, such as PFENet \cite{zhou2022extract} and PANet \cite{wang2019panet}, extract average prototypes of embeddings for each semantic category. Feature-matching methods, like CyCTR \cite{zhang2021few} and HDMNet \cite{peng2023hierarchical}, use pixel-level correlations between reference and target images to boost segmentation performance. In recent segmentation advancements, GFSS \cite{hossain2024visual} uses learned visual prompts with a transformer decoder to extract prototypes for generalized few-shot segmentation. We employ Stable Diffusion~\cite{rombach2022high} for automatic visual prompt generation aligned with text descriptions, enabling CLIP to produce rich semantic cost volumes through dual-prompt integration.

%% file: sec/3_Discussion.tex
\section{Discussion and Motivation}
\label{discussion}


In this section, we conduct experiments to analyze the modality gap \cite{jiang2023understanding,liang2022mind} between the aligned image embeddings and text embeddings extracted by pre-trained vision-language models, such as CLIP, as described in \secref{sec:domain-gap}. Additionally, we explore cost volume visualization to highlight the benefits of visual prompts by comparing the cost volumes generated from text and visual prompts, as shown in \secref{sec:cost-volume}. Based on the above analysis and observations, we propose combining visual and text prompts to enhance segmentation accuracy, particularly in scenarios involving both seen and unseen categories.

\subsection{Modality Gap of the Aligned Image and Text Embeddings}
\label{sec:domain-gap}

\begin{figure}[bt]

\centering
\includegraphics[width=1\linewidth]{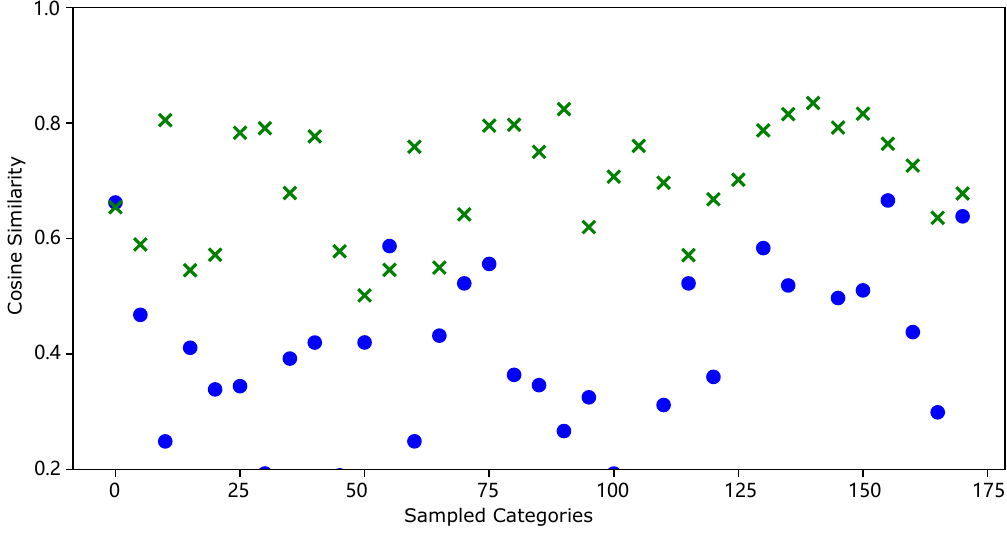}
\vspace{-20pt}
\caption{Visualization of cosine similarities comparing image embeddings $\mathbf{E}$ with text prompt embeddings  $\mathbf{T}$ (blue dots) and visual prompt embeddings  $\mathbf{V}$ (green crosses) across sampled categories.}
\vspace{-10pt}
\label{scores}
\end{figure}

To demonstrate the modality gap \cite{liang2022mind,jiang2023understanding} between aligned image and text embeddings, we conduct the experiment by following these steps: \ding{182} Given an image $\mathbf{I}$ and its corresponding text prompt $\mathcal{P}_t$, we use the pre-trained CLIP model to extract their embeddings $\mathbf{E}$ and $\mathbf{T}$, respectively. \ding{183} Using the pre-trained Stable Diffusion model \cite{rombach2022high} to generate the synthetic visual prompt $\mathcal{P}_v$ by inputting $\mathcal{P}_t$ and then extract its corresponding embedding $\mathbf{V}$ using the same CLIP image encoder. \ding{184} Calculating the cosine similarity between $\mathbf{E}$ and $\mathbf{T}$, as well as between $\mathbf{E}$ and $\mathbf{V}$, across more than one hundred samples.
As shown in \figref{scores}, the similarity between $\mathbf{E}$ and $\mathbf{V}$ is significantly higher than that between $\mathbf{E}$ and $\mathbf{T}$, demonstrating that the visual prompt $\mathcal{P}_v$, which shares the same modality as $\mathbf{I}$, can significantly alleviate the embedding modality gap.

\subsection{Effectiveness of the Visual Prompt in Semantic Segmentation}
\label{sec:cost-volume}

\begin{figure}
    \centering
    \begin{subfigure}{0.25\linewidth}
        \centering
        \includegraphics[width=1.0\linewidth]{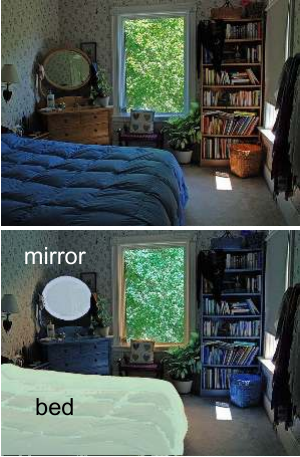}
        \caption{Image\&GT}
        \label{img}
    \end{subfigure}%
    \hfill
    \begin{subfigure}{0.25\linewidth}
        \centering
        \includegraphics[width=1.0\linewidth]{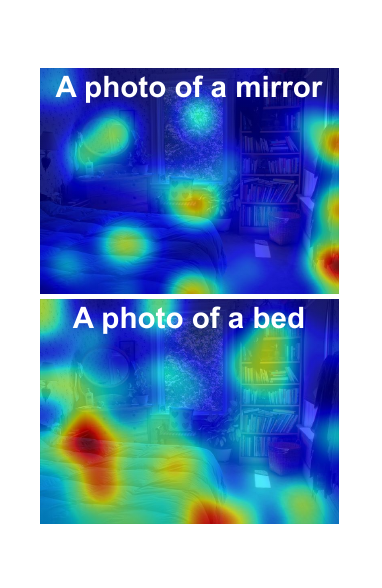}
        \caption{Text Prompt}
        \label{text_prompt}
    \end{subfigure}%
    \hfill
    \begin{subfigure}{0.25\linewidth}
        \centering
        \includegraphics[width=1.0\linewidth]{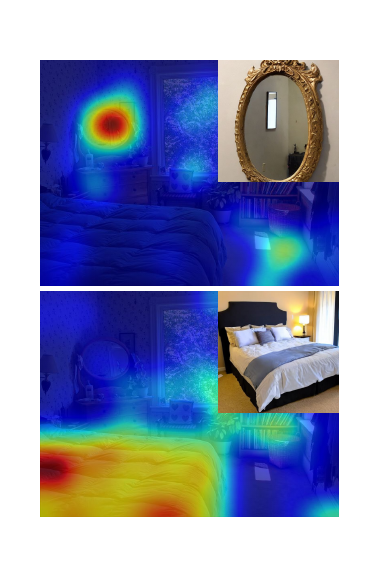}
        \caption{Visual Prompt}
        \label{visual}
    \end{subfigure}%
    \hfill
    \begin{subfigure}{0.25\linewidth}
        \centering
        \includegraphics[width=1.0\linewidth]{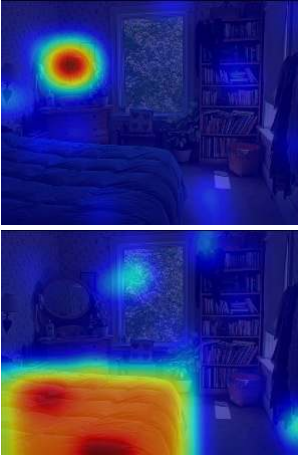}
        \caption{Ours}
        \label{Ours}
    \end{subfigure}
    \vspace{-20pt}
    \caption{Visualization of cost volume: (a) image with `mirror' and `bed' segments; (b) cost volume with text prompts; (c) cost volume with visual prompts; (d) aggregated cost volume. The top row represents the unseen class `mirror,' and the bottom row represents the seen class `bed'.}
    \label{prompt_visualization}
    \vspace{-15pt}
\end{figure}

\begin{figure*}[t]
\centering
    {\includegraphics[width=1.0\linewidth]{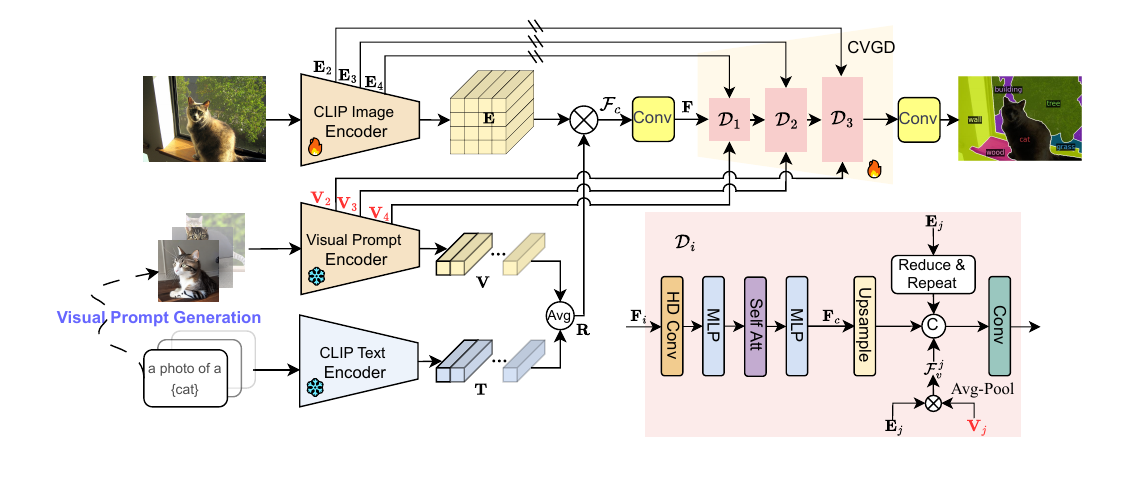}}
    \vspace{-15pt}
    \caption{Architecture of our DPSeg network. We begin by generating visual prompts based on text prompt templates, which are then combined with text prompts to create the dual-prompt cost volume. Subsequently, we incorporate multi-scale image features and corresponding visual prompt embeddings into the cost volume-guided decoder (CVGD) in a progressive manner.}
    \label{pipeline}
    \vspace{-15pt}
\end{figure*}

To analyze the effectiveness of the visual prompt in semantic segmentation, we conduct the experiment as follows: \ding{182} Using the pre-trained CLIP image encoder, we extract the image feature from the layer before pooling. \ding{183} Using the same approach as in \secref{sec:domain-gap}, we extract the visual prompt embedding $\mathbf{V}$ and the text prompt embedding $\mathbf{T}$. \ding{184} We then calculate the similarity between the image feature of each pixel location and the visual prompt embedding $\mathbf{V}$ as well as the text prompt embedding $\mathbf{T}$. As shown in the heatmap in \figref{prompt_visualization}, visual prompts yield sharper and more accurate semantic contours than text prompts, particularly for details like ``mirror'' and ``bed'' in \figref{prompt_visualization} (b) and (c).

Inspired by this observation, in this work we propose to leverage both visual and text prompts to enhance the segmentation performance. Specifically, we average the visual and text prompt embeddings and calculate the cosine similarity as described above. As illustrated in \figref{prompt_visualization} (d), using both visual and text prompts reduces noise and captures more accurate semantic information.

%% file: sec/4_Method.tex
\section{Method}
Building on insights from \secref{discussion}, we propose our novel framework \textbf{DPSeg} for open-vocabulary semantic segmentation, comprising three core components: a dual-prompt cost volume generation module (see \secref{Dual-Modal}), a cost volume-guided decoder (see \secref{decoder}), and a semantic-guided prompt refinement module (see \secref{refinement}), as shown in \figref{pipeline} and \figref{infer}. The dual-prompt cost volume generation module utilizes both text and visual prompts to generate a pixel-level cost volume $\mathcal{F}_c$ with image features. The cost volume-guided decoder integrates multi-scale cost volumes $\mathcal{F}_v^2,\mathcal{F}_v^3, \mathcal{F}_v^4$ from the image and visual prompts to guide segmentation map prediction. To address misalignment and uncertainty in visual prompt generation, we propose a two-pass inference strategy, semantic-guided prompt refinement, to enhance alignment and improve segmentation accuracy.

\subsection{Dual-Prompt Cost Volume Generation}
\label{Dual-Modal}
Prior open-vocabulary semantic segmentation approaches \cite{cho2024cat,xie2024sed} mainly rely on text-image feature alignment using CLIP to generate pixel-level cost volumes. However, such cross-modal alignment can be limited in precision due to inherent differences between text and image representations, even with advanced pre-training as discussed in \secref{discussion}. To address this, we introduce a cross-modal text-visual prompt cost volume generation, which enhances the alignment within the visual domain by introducing higher intra-modality similarity.

\textbf{Image encoder.} For an input RGB image $\mathbf{I}$, we employ a hierarchical encoder to extract image features \(\mathbf{E}_j, j \in (2,3,4,5)\) at progressively coarser resolutions, corresponding to scales of \(4 \times\), \(8 \times\), \(16 \times\), and \(32 \times\) smaller than the original input size. The final layer, \(\mathbf{E}_5\), is processed through an MLP layer to align the dimensionality with the prompt embeddings, yielding the final image embedding \(\mathbf{E} \in \mathbb{R}^{H \times W \times D_z}\), where $H, W,$ and $D_z$ represent the height, width, and number of channels respectively.

\textbf{Text prompt embedding generation.} For generating text prompt embeddings, we adopt a diverse set of descriptive templates following the strategies in \cite{cho2024cat,xie2024sed,liang2023open,gu2021open}. These templates provide varied descriptions for each category \( C_k \), such as ``a photo of a \{$C_k$\}'', ``a \{$C_k$\} in the scene'', etc. Each description \( P_t^{(k,m)} \) is processed through CLIP's text encoder, producing text embeddings $\mathbf{T} = \{ \mathbf{T}_{k,m} \mid k = 1, \ldots, K; \, m = 1, \ldots, M \} \in \mathbb{R}^{K \times M \times D_z}$, where \( K \) is the total number of categories, M is the number of templates per category, and \( D_z \) is the dimension of the text prompt embeddings, which matches that of the image embeddings for consistency. This variety of templates enhances the semantic richness of each category representation, supporting robust category detection across diverse contexts.

\textbf{Visual prompt embedding generation.} For visual prompt generation, we employ a pre-trained Stable Diffusion model \cite{rombach2022high} to generate visual prompts by inputting text prompt $\mathcal{P}_t^{(k,m)}$, with each template producing a distinct prompt prototype for the category. The generated visual prompts are processed through a visual prompt encoder, structurally identical to the image encoder. The visual prompt encoder produces an embedding \( \mathbf{V} \in \mathbb{R}^{K \times M \times D_z} \), aligned with the text prompt embeddings for seamless integration. Additionally, we extract multi-scale features $\mathbf{V}_j \in \mathbb{R}^{H_j \times W_j \times K \times M \times D_j}$ (for \(j = 2, 3, 4\)), where \(H_j\) and \(W_j\) denote the spatial dimensions downsampled by 4, 8 and 16, respectively; \(K\) and \(M\) represent the category and prototypes of visual prompts, and \(D_j\) is the feature dimension at each level \(j\). These multi-scale features further enrich the cost volume, providing fine-grained semantic information that enhances decoding in the segmentation process.

\textbf{Combined Cost Volume Generation.} To utilize both visual and text prompt embeddings, we average them to create a unified representation $\mathbf{R}$, denoted as 
\[
\mathbf{R} = \text{Avg}(\mathbf{V} + \mathbf{T}).
\]
Then we calculate the dual-prompt cost volume \(\mathcal{F}_c \in \mathbb{R}^{H \times W \times K \times M}\) using the \(\mathbf{R}\) and image feature $\mathbf{E}$, as follows:

\begin{equation}
    \mathcal{F}_c(x, y, k, m) = \frac{\mathbf{E}(x, y) \cdot \mathbf{R}(k, m)}{\|\mathbf{E}(x, y)\| \|\mathbf{R}(k, m)\|},
\end{equation}
where \((x, y)\) represents the spatial location, and \((k, m)\) denotes the category and template index, respectively. To facilitate efficient processing of this high-dimensional cost volume, we apply a convolution layer independently to each cost slice, producing an initial cost volume embedding. This embedding is then fed into the decoder as $\mathbf{F} \in \mathbb{R}^{(H \times W) \times K \times d_F}$, where \(d_F\) is the cost volume embedding dimension. 

\subsection{Cost Volume-Guided Decoder}
\label{decoder}

Building on the findings in \secref{discussion}, we design a Cost Volume-Guided Decoder that leverages dual-prompt cost volume and hierarchical encoder structure. At each decoder stage $\mathcal{D}_i$ ($i \in (1,2,3)$), the input feature map $\mathbf{F}_i$ is processed through a hybrid dilated convolution (HD Conv) module with a 3×3 kernel and dilation rates of 1, 2, and 4, providing large receptive fields (3×3, 5×5, and 9×9) with fewer parameters than standard large-kernel convolutions. These features are merged via element-wise addition, followed by normalization, an MLP with GeLU activation, and a self-attention layer for category-level aggregation.

To integrate multi-scale image features from the encoder, which captures rich local details, we first upsample the feature map $\mathbf{F}_c$ following the self-attention layer by a factor of 2 using deconvolution. Each scale feature $\mathbf{E}_j, j \in (2,3,4)$ is then reduced by a factor of 16 through convolution, repeated the number of categories N times to align with the category dimension of the upsampled feature map, and finally concatenated with it. As suggested in \cite{cho2024cat}, we avoid back-propagating these intermediate features directly to the image encoder to prevent potential performance degradation.

Instead of upsampling initial cost volume, we employ intra-modal similarity by integrating intermediate-scale image features \(\mathbf{E}_j \in \mathbb{R}^{ K \times M \times D_j},j \in (2,3,4)\), with corresponding multi-scale visual prompt features \(\mathbf{V}_j \in \mathbb{R}^{H_j \times W_j \times K \times M \times D_j}, j \in (2,3,4)\) to compute the visual cost volumes, thereby enhancing semantic coherence across scales. This method captures fine-grained semantic information at multiple levels, yielding visual cost volumes $\mathcal{F}_v^j$ at each stage, calculated as:
\begin{equation}
    \mathcal{F}_v^j(x, y, k,m) = \frac{Pool(\mathbf{V}_j(k, m)) \cdot \mathbf{E}_j(x, y)}{\|Pool(\mathbf{V}_j(k, m))\| \|\mathbf{E}_j(x, y)\|}.
\end{equation}
Here, global average pooling is applied along the $H_j$ and $W_j$. Each channel in the visual cost volume directly aligns with the corresponding decoding feature channels at each decoder stage, eliminating the need for additional upsampling and preserving both semantic and spatial details. The final high-dimensional feature map is then mapped to the segmentation output through a convolution layer, enabling accurate predictions across all semantic categories for open-vocabulary semantic segmentation.

\begin{figure}[bt]

\centering
\includegraphics[width=1\linewidth]{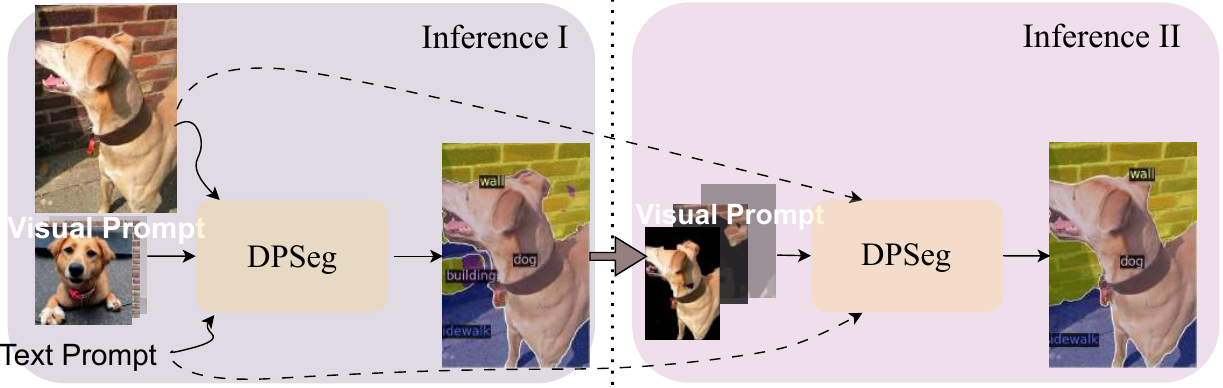}
\vspace{-15pt}
\caption{Structure of Semantic-Guided Prompt Refinement.}
\vspace{-15pt}
\label{infer}
\end{figure}

\begin{table*}[t]
\centering
\begin{adjustbox}{width=\textwidth}
\renewcommand{\arraystretch}{0.85}
\begin{tabular}{l|cc|c|cccccc}
\toprule
Method & VLM & Backbone & Training Dataset & A-847 & PC-459 & A-150 & PC-59 & PAS-20 \\
\midrule
\midrule
ZS3Net \cite{bucher2019zero}& - & ResNet-101 & PASCAL VOC & - & - & - & 19.4 & 38.3 \\
LSeg \cite{xian2019semantic}& ViT-B/32 & ResNet-101 & PASCAL VOC-15 & - & - & - & - & 47.4 \\
LSeg+ \cite{ghiasi2022scaling}& ALIGN RN-101 & ResNet-101 & COCO-Stuff & 2.5 & 5.2 & 13.0 & 36.0 & - \\
Han et al. \cite{han2023global}& ViT-B/16 & ResNet-101 & COCO Panoptic \cite{kirillov2019panoptic}& 3.5 & 7.1 & 18.8 & 45.2 & 83.2 \\
ZegFormer \cite{ding2022decoupling}& ViT-B/16 & ResNet-101 & COCO-Stuff & 5.6 & 10.4 & 18.0 & 45.5 & 89.5 \\
SimSeg \cite{xu2022simple}& ViT-B/16 & ResNet-101 & COCO-Stuff & 7.0 & - & 20.5 & 47.7 & 88.4 \\
OpenSeg \cite{ghiasi2022scaling}& ALIGN & ResNet-101 & COCO Panoptic \cite{kirillov2019panoptic} + Loc. Narr. \cite{pont2020connecting}& 4.4 & 7.9 & 17.5 & 40.1 & - \\
PACL \cite{mukhoti2023open}& ViT-B/16 & - & GCC \cite{sharma2018conceptual} + YFCC\cite{thomee2016yfcc100m} & - & - & 31.4 & 50.1 & 72.3 \\
OVSeg \cite{liang2023open}& ViT-B/16 & ResNet-101c & COCO-Stuff+COCO Caption & 7.1 & 11.0 & 24.8 & 53.3 & 92.6 \\
CAT-Seg \cite{cho2024cat}& ViT-B/16 & ResNet-101 & COCO-Stuff & 8.4 & 16.6 & 27.2 & 57.5 & 93.7 \\
SAN \cite{xu2023side}& ViT-B/16 & - & COCO-Stuff & 10.1 & 12.6 & 27.5 & 53.8 & 94.0 \\
EBSeg \cite{shan2024open}& ViT-B/16 & - & COCO-Stuff & 11.1 & 17.3 & 30.0 & 56.7 & 94.6 \\
SCAN \cite{liu2024open} & ViT-B/16 & Swin-B & COCO-Stuff & 10.8 & 13.2 & 30.8 & \textbf{58.4} & \textbf{97.0} \\ 
SED \cite{xie2024sed}& ConvNeXt-B & - & COCO-Stuff & 11.4 & 18.6 & 31.6 & 57.3 & 94.4 \\
\rowcolor{lightgray}DPSeg (Inference I) & ConvNeXt-B & - & COCO-Stuff & \underline{12.0} & \underline{19.5} & \underline{32.9} & \underline{58.1} & 96.0 \\
\rowcolor{lightgray}DPSeg (Inference II, \textbf{Ours}) & ConvNeXt-B & - & COCO-Stuff & \textbf{12.5} & \textbf{20.1} & \textbf{33.3} & \textbf{58.4} & \underline{96.9} \\
\midrule
LSeg \cite{xian2019semantic}& ViT-B/32 & ViT-L/16 & PASCAL VOC-15 & - & - & - & - & 52.3 \\
OpenSeg \cite{ghiasi2022scaling}& ALIGN & Eff-B7 \cite{tan2019efficientnet}& COCO Panoptic \cite{kirillov2019panoptic} + LOc. Narr. \cite{pont2020connecting}& 8.1 & 11.5 & 26.4 & 44.8 & - \\
OVSeg \cite{liang2023open}& ViT-L/14 & Swin-B & COCO-Stuff+COCO Caption & 9.0 & 12.4 & 29.6 & 55.7 & 94.5 \\
Ding et al. & ViT-L/14 & - & COCO Panoptic \cite{kirillov2019panoptic}& 8.2 & 10.0 & 23.7 & 45.9 & - \\
ODISE \cite{xu2023open} & ViT-L/14 & - & COCO Panoptic \cite{kirillov2019panoptic}& 11.1 & 14.5 & 29.9 & 57.3 & - \\
HIPIE \cite{wang2024hierarchical} & BERT-B & ViT-H & COCO Panoptic \cite{kirillov2019panoptic}& - & - & 29.0 & 59.3 & - \\
SAN \cite{xu2023side}& ViT-L/14 & - & COCO-Stuff & 13.7 & 17.1 & 33.3 & 60.2 & 95.5 \\
CAT-Seg \cite{cho2024cat}& ViT-L/14 & Swin-B & COCO-Stuff & 10.8 & 20.4 & 31.5 & \underline{62.0} & 96.6 \\
FC-CLIP \cite{yu2023convolutions}& ConvNeXt-L & - & COCO Panoptic \cite{kirillov2019panoptic}& 14.8 & 18.2 & 34.1 & 58.4 & 95.4 \\
EBSeg \cite{shan2024open}& ViT-L/14 & - & COCO-Stuff & 13.7 & 21.0 & 32.8 & 60.2 & 96.4 \\
SCAN \cite{liu2023open} & ViT-L/14 & Swin-B & COCO-Stuff & 14.0 & 16.7 & 33.5 & 59.3 & 97.2 \\
USE+SAM \cite{10656965}& ViT-L/14 & Swin-B & COCO-Stuff & 13.3 & 14.7 & \underline{37.0} & 57.8 & - \\
SED \cite{xie2024sed}& ConvNeXt-L & - & COCO-Stuff & 13.9 & 22.6 & 35.2 & 60.6 & 96.1 \\
\rowcolor{lightgray}DPSeg (Inference I) & ConvNeXt-L  & - & COCO-Stuff & \underline{14.9} & \underline{23.5} & 36.4 & \underline{62.0} & \underline{97.4} \\
\rowcolor{lightgray}DPSeg (Inference II, \textbf{Ours}) & ConvNeXt-L & - & COCO-Stuff & \textbf{15.7} & \textbf{24.1} & \textbf{37.1} & \textbf{62.3} & \textbf{98.5} \\
\bottomrule
\end{tabular}
\end{adjustbox}
\vspace{-5pt}
\caption{Comparison with state-of-the-art methods on five open-vocabulary semantic segmentation test sets. mIoU results are shown, with the best in bold and the second-best underlined. In both configurations of the VLM model, our DPSeg under Inference I and II achieves superior performance, with Inference II consistently outperforming all other SOTA methods.}
\label{quantitative}
\vspace{-12pt}
\end{table*}

\subsection{Semantic-Guided Prompt Refinement }
\label{refinement}
The visual prompt generation process during training and initial inference introduces variability, as visual prompts generated from text prompts may not fully align with the input image $\mathbf{I}$. This misalignment can impact segmentation accuracy, particularly for fine-grained details or objects that are not well-represented by the initial visual prompt. 
To address this challenge, we propose a semantic-guided prompt refinement inference strategy, illustrated in \figref{infer}. The first inference (Inference I) pass employs text prompts and initial visual prompts to produce preliminary segmentation results. For each category detected in the first pass, its corresponding segmentation mask is used to crop the objects, which will replace all visual prompts of that category in the second inference pass (Inference II) while maintaining initial prompts for undetected categories and text prompts. This instance-aware refinement enhances the cost volume's semantic alignment with actual objects in the current image, achieving more precise object boundaries and local details.

%% file: sec/5_Experiment.tex
\begin{figure*}[t]
\centering
    {\includegraphics[width=1.0\linewidth]{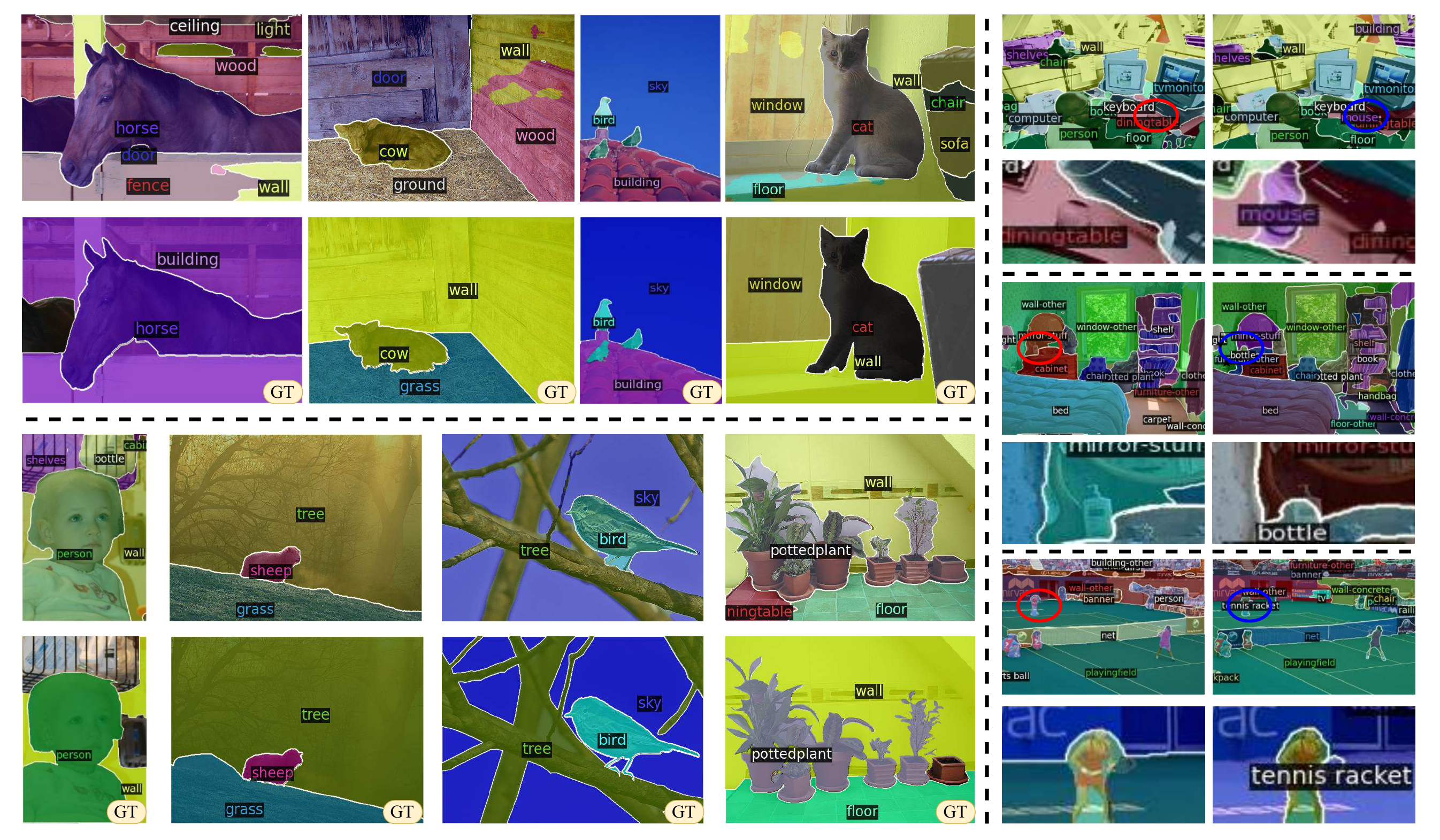}}
    \vspace{-15pt}
    \caption{\textit{Visualization examples of our model on open vocabulary semantic segmentation.} \textbf{First four columns}: We present results from our model alongside the corresponding ground truth. \textbf{Fifth and sixth columns}: SED \cite{xie2024sed} results are shown for comparison, highlighting our model’s improved segmentation of small foreground objects, such as the mouse, bottle, and tennis racket. }
    \label{qualitative}
    \vspace{-10pt}
\end{figure*}

\section{Experiments}

\subsection{Setups}
{\bf Datasets.} To acquire object information and evaluate our models, we follow a dataset setup similar to that in \cite{xie2024sed}. We train our model on the \textbf{COCO-Stuff} \cite{caesar2018cvpr} dataset and evaluate its performance across five additional benchmarks: \textbf{Pascal VOC} \cite{everingham2011pascal}, \textbf{Pascal Context} \cite{mottaghi2014role} (referred to as \textit{PC-59} and \textit{PC-459}), \textbf{ADE20K-150} ( referred to as \textit{A-150}) \cite{zhou2017scene}, and \textbf{ADE20K-847} (referred to as \textit{A-847}) \cite{zhou2017scene}. \textbf{COCO-Stuff} is a large-scale semantic segmentation dataset comprising 171 densely annotated classes, with 118\textit{k} training images, 5\textit{k} validation images, and 41\textit{k} test images. \textbf{Pascal VOC}, a traditional dataset for segmentation and object detection, includes 20 classes, with 1464 training and 1449 validation images. \textbf{Pascal Context} is an extension of \textbf{Pascal VOC}, with 1.5\textit{k} training images and 1.5\textit{k} validation images, offering two types of annotations: \textit{PC-59} (59 classes) and \textit{PC-459} (459 classes). The \textbf{ADE20K} dataset consists of 20\textit{k} training images and 2\textit{k} validation images and is split into two test sets: \textit{A-150} with 150 annotated classes, and \textit{A-847} with 847 common semantic categories.

{\bf Metrics.} We evaluate our model using Intersection over Union (IoU). Our model is evaluated without resorting to finetuning in various settings, demonstrating the enhancements it achieves.

\subsection{Implementation Details}
\textbf{Network Architectures.} We employ the OpenAI pre-trained CLIP model \cite{radford2021learning} as the backbone, using either ConvNeXt-B or ConvNeXt-L as the image encoder and visual prompt encoder. For the text encoder, the text embedding dimension \(D_z\) is set to 640 for ConvNeXt-B and 768 for ConvNeXt-L. The initial cost volume fed into the decoder is set to a dimension of 128, with the output hidden dimensions of each decoder stage as [62, 32, 16].
For consistency, visual prompts are generated at 768$\times$768 resolution, matching the input image size during training and inference. To improve training efficiency, the text and visual prompt encoders are frozen, with only the image encoder and cost volume-guided decoder being trained. Training uses a per-pixel binary cross-entropy loss.
Our models are implemented in PyTorch \cite{paszke2019pytorch} and Detectron2 \cite{wu2019detectron2}. We employ the AdamW optimizer with an initial learning rate of \(2 \times 10^{-4}\), weight decay of \(1 \times 10^{-4}\), and an additional factor \(\alpha=0.01\)  to mitigate overfitting. The mini-batch size is set to 4, and models are trained for 80\textit{k} iterations on two V100 GPUs.

\subsection{Quantitative Results}

As shown in \tableref{quantitative}, in the base-VLM configuration, our approach achieves significant gains over state-of-the-art methods like SED \cite{xie2024sed}, with improvements of +1.1\%, +1.5\%, +2.5\%, and +1.1\% mIoU on the \textit{A-187}, \textit{PC-459}, \textit{A-150}, and \textit{PC-59} datasets, respectively. Between our two inference stages, Inference II shows an additional gain of +0.87\% mIoU over Inference I. In the \textit{Large}-scale VLM configuration, our method further outperforms SED, achieving gains of +1.8\%, +1.5\%, +1.9\%, +1.7\%, and +2.4\% mIoU across all datasets, with Inference II yielding an extra +0.92\% mIoU over Inference I. These results underscore the effectiveness of our proposed method in advancing OVSS.

\subsection{Qualitative Results}

\begin{figure}
    \centering
    \begin{subfigure}{0.49\linewidth}
        \centering
        \includegraphics[width=1.0\linewidth]{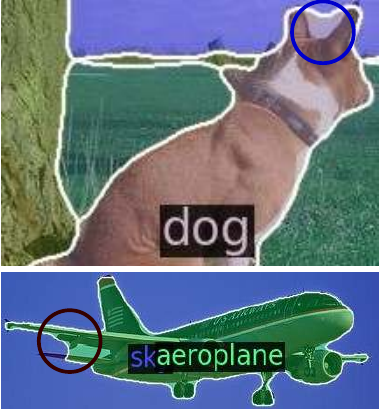}
        \caption{Inference I}
        \label{infer1}
    \end{subfigure}%
    \hfill
    \begin{subfigure}{0.49\linewidth}
        \centering
        \includegraphics[width=1.0\linewidth]{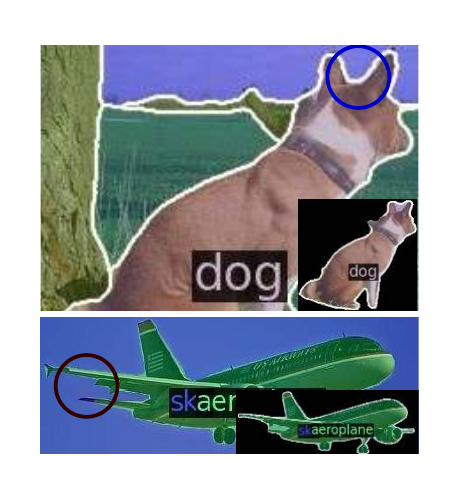}
        \caption{Inference II}
        \label{inference}
    \end{subfigure}%
    \vspace{-5pt}
    \caption{Comparison between inference I and II with our refinement strategy.}
    \vspace{-20pt}
    \label{sgpr}
\end{figure}

\figref{qualitative} shows segmentation examples from our model. In the first four columns, we display samples from the ADE-150 and PC-59 datasets alongside the ground truth, illustrating our model’s strong performance across diverse scenarios. Our segmentation results not only capture large foreground objects in both seen and unseen classes but also detect additional categories not annotated in the ground truth, such as ``\textit{wood},'' ``\textit{fence},'' and ``\textit{chair}'' in the first, second, and fourth images of the first row.
We also compare three cases between SED \cite{xie2024sed} and our model. As shown in the last two columns of \figref{qualitative}, our method accurately segments smaller objects near larger ones, such as the mouse, bottle, and tennis racket, effectively preserving semantic information across scales.
In \figref{sgpr}, we compare two cases of our model between Inference I and Inference II. The results demonstrate our method's effectiveness in enhancing semantic alignment and capturing finer segmentation details based on the initial segmentation output.

\subsection{Ablation Study}
For simplicity, all ablation studies are conducted using the strategy and results from Inference I.

{\bf Prompt strategy and cost volume generation.}
In this section, we conduct ablation studies to explore how different prompt strategies and cost volume generation strategies affect the segmentation results.
From \tableref{prompt} (a), we observe that the visual-only prompt outperforms the text-only prompt. By averaging the visual and text embedding, achieving better overall performance. In \tableref{prompt} (b), we explore two additional cost volume generation methods: \ding{182} generating separate cost volumes for text and visual prompts, then concatenating them, and \ding{183} directly averaging the two cost volumes. The results clearly show that our approach—averaging the embeddings of both modalities before calculating cosine similarity—yields superior performance.

\begin{table}[t]
\centering
 \renewcommand{\arraystretch}{1.5} 
\begin{adjustbox}{width=0.95\linewidth}
\begin{tabular}{c|c|c|c|c|c|c}
\toprule
\multirow{4}{*}{(a)} & \textbf{Prompt Strategy} & A-847 & PC-459 & A-150 & PC-59 & PAS-20 \\ \hhline{~======}
& $\mathbf{T}$ & 10.4 & 17.4 & 30.6 & 56.2 & 93.4 \\
& $\mathbf{V}$ & 11.1 & 18.0 & 31.2 & 56.9 & 94.5 \\\rowcolor{lightgray}
& $\mathrm{Avg}(\mathbf{T},\mathbf{V})$ (\textbf{Ours}) & 12.0 & 19.5 & 32.9 & 58.1 & 96.0 \\
\midrule
\multirow{4}{*}{(b)} & \textbf{Cost Volume Generation} & A-847 & PC-459 & A-150 & PC-59 & PAS-20 \\ \hhline{~======}
 & Cat(cos($\mathbf{T,E}$),cos($\mathbf{V,E})$) & 11.7 & 19.2 & 32.3 & 57.8 & 95.4 \\
& Avg(cos($\mathbf{T,E}$),cos($\mathbf{V,E})$) & 11.9 & 19.4 & 32.7 & 58.0 & 95.7 \\\rowcolor{lightgray}
& $\mathcal{F}_c$ (\textbf{Ours}) & 12.0 & 19.5 & 32.9 & 58.1 & 96.0 \\
\bottomrule
\end{tabular}
\end{adjustbox}
\caption{\textbf{Ablation study on prompt types and cost volume generation strategies.} Results are shown for text-only ($\mathbf{T}$), visual-only ($\mathbf{V}$), and combined text-visual prompt embeddings, along with comparisons of three cost volume generation solutions}
\label{prompt}
\vspace{-7pt}
\end{table}

\begin{table}[t]
\centering
\renewcommand{\arraystretch}{1.5}  
\begin{adjustbox}{width=1\linewidth}  
\begin{tabular}{c|ccccc}
\toprule
\textbf{Fusion Strategy}  & A-847 & PC-459 & A-150 & PC-59 & PAS-20 \\
\hline
\hline
    $\mathcal{F}_c$ & 10.6 & 17.9 & 31.6 & 57.0 & 94.9 \\
    $\mathcal{F}_c, \mathcal{F}_v^2$ & 11.4 & 18.8 & 32.1 & 57.6 & 95.5\\
    $\mathcal{F}_c, \mathcal{F}_v^{2,3} $ & 11.7 & 19.2 & 32.5 & 57.9 & 95.7 \\
    \hline
    \rowcolor{lightgray}
    $\mathcal{F}_c, \mathcal{F}_v^{2,3,4} $ (\textbf{Ours}) & 12.0 & 19.5 & 32.9 & 58.1 & 96.0  \\
        $\mathcal{F}_c, \mathcal{F}_c (2\times,4\times,8\times) $ & 11.4 & 19.0 & 32.4 & 58.0 & 95.6 \\
\bottomrule
\end{tabular}
\end{adjustbox}
\vspace{-5pt}
\caption{\textbf{Ablation study on cost volume guidance strategies.} $\mathcal{F}_c$ denotes the cost volume in our model, while $\mathcal{F}_v^{i}$ represents cost volume at layer $i$, calculated from different levels of the image features and visual prompts. $\mathcal{F}_c(2\times)$ refers to the method where the cost volume $\mathcal{F}_c$ is upsampled and concatenated.}
\label{tab:cost_volume_fusion}
\vspace{-5pt}
\end{table}

\begin{figure}
    \centering
    \begin{subfigure}{0.33\linewidth}
        \centering
        \includegraphics[width=1.0\linewidth]{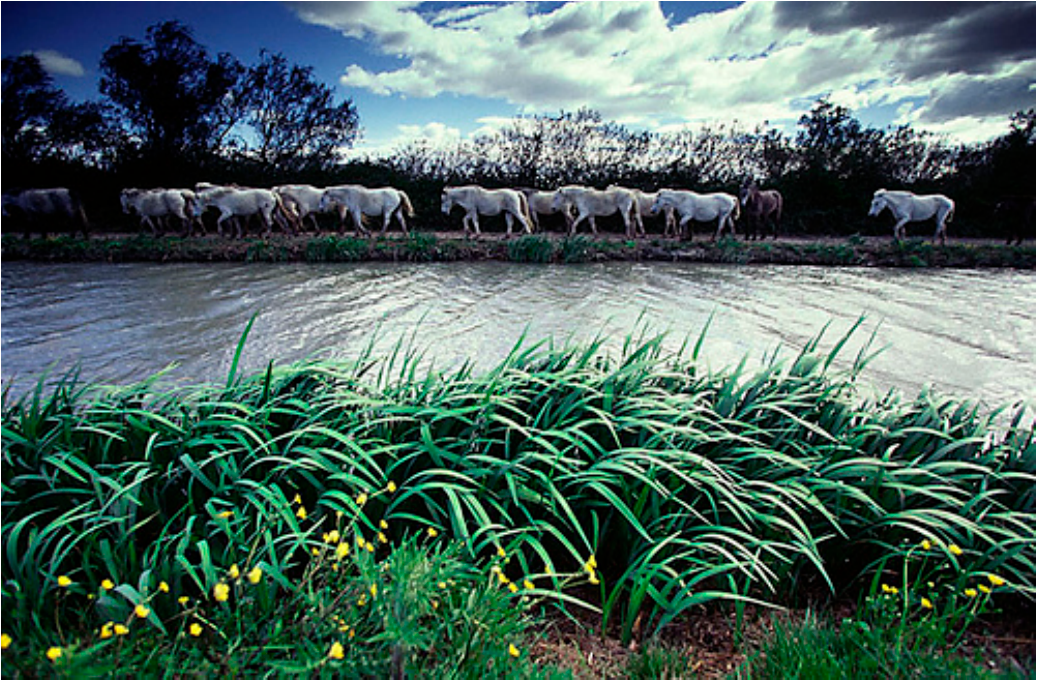}
        \caption{Image}
        \label{Text}
    \end{subfigure}%
    \hfill
    \begin{subfigure}{0.33\linewidth}
        \centering
        \includegraphics[width=1.0\linewidth]{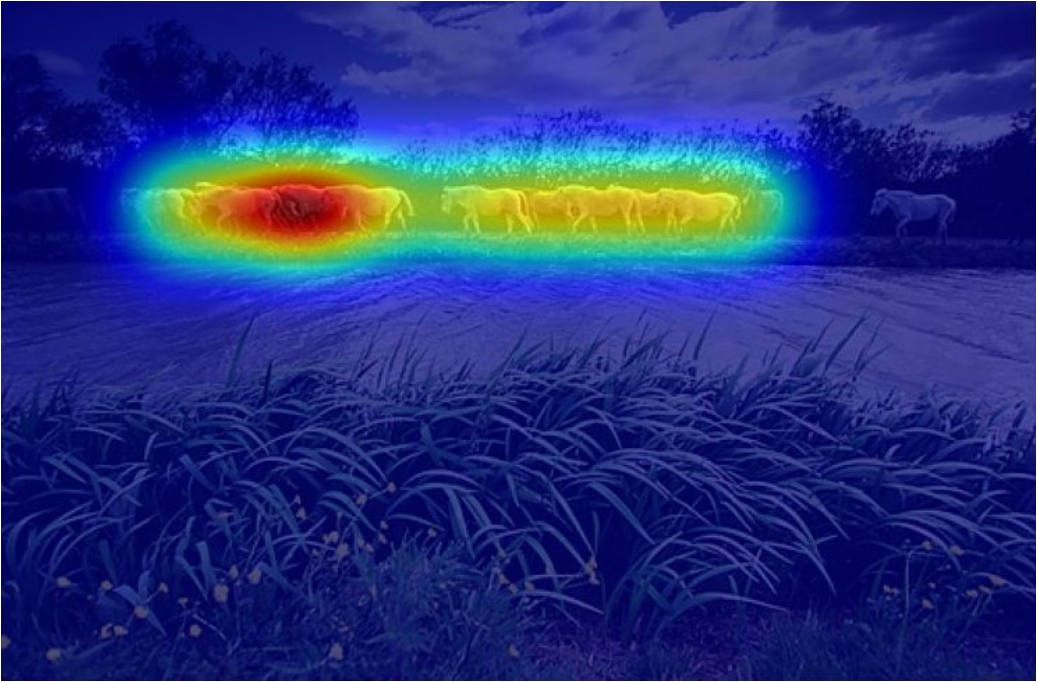}
        \caption{$\mathcal{F}_c$ (8$\times$)}
        \label{Visual}
    \end{subfigure}%
    \hfill
    \begin{subfigure}{0.33\linewidth}
        \centering
        \includegraphics[width=1.0\linewidth]{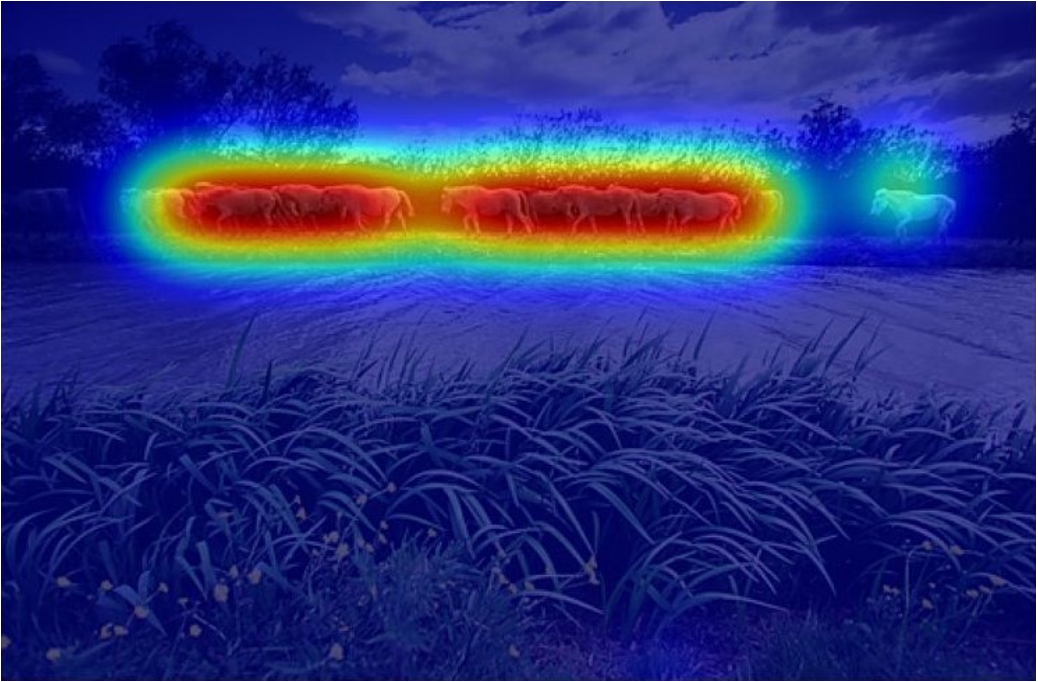}
        \caption{ $\mathcal{F}_v^{4}$}
        \label{Ours_horse}
    \end{subfigure}
    \vspace{-5pt}
    \caption{\textbf{Comparison of the visualized cost volumes.} Evaluating the effects of cost volume by comparing the (b) 8$\times$ upsampled cost volume with the (c) the cost volume derived from 4th intermediate feature and visual prompt.}
    \label{horse}
    \vspace{-15pt}
\end{figure}

\textbf{Cost volume-guided decoder.} We examine the effect of various cost volume guidance strategies in our decoder, as shown in \tableref{tab:cost_volume_fusion}. When using only the cost volume \(\mathcal{F}_c\) generated by text and visual embeddings, segmentation performance is lowest (first row). Performance steadily improves as we progressively integrate multi-scale visual cost volumes \(\mathcal{F}_v^{2,3,4}\), achieving an average increase of 1.8\% mIoU across all datasets when all scales are incorporated (second and third rows). Replacing \(\mathcal{F}_v^{2,3,4}\) with an upsampled \(\mathcal{F}_c\) results in a performance drop (last row).
\figref{horse} illustrates how our multi-scale cost volume captures finer semantic details, such as a single horse standing apart from a crowd—details that upsampled cost volumes fail to capture. This progressive integration of multi-scale cost volumes preserves spatial details, enhances multi-scale feature utilization, and maintains semantic consistency across scales, enabling our model to capture fine details for improving segmentation accuracy.

\textbf{Number of Templates.}
\tableref{templates} shows the results of an ablation study examining the impact of the number of templates on segmentation performance. In the experiment, each text template is paired with a corresponding visual prompt prototype. As the number of templates increases, we observe a significant improvement in performance. This result suggests that a diverse set of templates better captures variations in real-world scenarios, as a single prompt may fail to represent all contexts. For example, `a photo of a bird' may not accurately depict an image with multiple birds, which would be better represented by `a photo of many birds'.

\begin{table}[t]
\centering
 \renewcommand{\arraystretch}{1.3} 
\begin{adjustbox}{width=0.95\linewidth}
\begin{tabular}{c|ccccc}
\toprule
 \textbf{Number of Templates} & A-847 & PC-459 & A-150 & PC-59 & PAS-20  \\
\hline
\hline
1 & 3.6 & 7.8 & 15.6 & 32.1 & 72.3  \\
10 & 6.9 & 11.2 & 20.3 & 41.3 & 81.7 \\
20 & 7.8 & 14.9 & 27.9 & 47.3 & 86.7 \\
40 & 10.1 & 16.8 & 29.6 & 55.9 & 91.8 \\ \rowcolor{lightgray}
80 & 12.0 & 19.5 & 32.9 & 58.1 & 96.0 \\
\bottomrule
\end{tabular}
\end{adjustbox}
\vspace{-5pt}
\caption{\textbf{Ablation study on the number of prompt templates.} Both text and visual prompts are used.}
\label{templates}
\vspace{-5pt}
\end{table}

\begin{table}[t]
\centering
 \renewcommand{\arraystretch}{1.3} 
\begin{adjustbox}{width=0.95\linewidth}
\begin{tabular}{c|ccccc}
\toprule
 \textbf{Visual Prompt Generation} & A-847 & PC-459 & A-150 & PC-59 & PAS-20  \\
\hline
(1) 20\% Gaussian Noise & 12.0 & 19.2 & 32.5 & 57.9 & 95.5  \\
(2) 60\% Gaussian Noise & 11.6 & 18.8 & 32.9 & 57.6 & 95.0 \\
(3) 80\% Gaussian Noise & 11.5 & 18.7 & 31.7 & 57.3 & 94.7 \\ 
\hline
(4) LayerDiffuse\cite{zhang2024transparentimagelayerdiffusion} & 12.1 & 19.4 & 32.7 & 58.0 & 95.8 \\
(5) SD-V1.5 & 12.0 & 19.5 & 32.9 & 58.1 & 96.0\\
(6) SD-V3\cite{esser2024scaling} & 12.2 & 19.6 & 32.8 & 58.3 & 96.3 \\
\hline
(7) SED (text prompt only) & 11.4 & 18.6 & 31.6 & 57.3 & 94.4 \\
\bottomrule
\end{tabular}
\end{adjustbox}

\vspace{-5pt}
\caption{\textbf{Ablation study on visual prompts generation}. Methods(1)–(3) use SD-v1.5 with varying Gaussian noise levels (e.g., 80\% means noise variance is 80\% of the maximum), methods(4)–(6) employ LayerDiffuse, SD-v1.5, and SD-v3, respectively, and method(7) shows SED segmentation performance.}
\label{generation}
\vspace{-17pt}
\end{table}

\textbf{Quality of visual prompts.} 
The quality of visual prompts generated by different generation models varies, as shown in \tableref{generation}. Our experimental results highlight the robustness of our approach: even when using low-quality visual prompts (e.g., 80\% Gaussian noise), our method consistently outperforms the text-prompt-only baseline (e.g., SED) in segmentation. Furthermore, employing different text-to-image generation models ((4)-(6) in \tableref{generation}) leads to minimal performance variation. These findings reinforce our commitment to integrating advanced generation techniques to enhance OVSS for both research and real-world applications.

%% file: sec/6_Conclusion.tex
\section{Conclusion}
In conclusion, we introduce \textbf{DPSeg}, a novel dual-prompt framework for open-vocabulary semantic segmentation (OVSS). Our approach synergistically combines text and visual prompts with multi-scale features, while introducing an innovative refinement strategy to enhance semantic alignment and fine-grained detail preservation. Extensive benchmarks validate its state-of-the-art performance, demonstrating the effectiveness of our dual-prompt cost volume learning approach.

%% file: main.bbl
\begin{thebibliography}{69}
\providecommand{\natexlab}[1]{#1}
\providecommand{\url}[1]{\texttt{#1}}
\expandafter\ifx\csname urlstyle\endcsname\relax
  \providecommand{\doi}[1]{doi: #1}\else
  \providecommand{\doi}{doi: \begingroup \urlstyle{rm}\Url}\fi

\bibitem[Abdelfattah et~al.(2023)Abdelfattah, Guo, Li, Wang, and Wang]{abdelfattah2023cdul}
Rabab Abdelfattah, Qing Guo, Xiaoguang Li, Xiaofeng Wang, and Song Wang.
\newblock Cdul: Clip-driven unsupervised learning for multi-label image classification.
\newblock In \emph{Proceedings of the IEEE/CVF international conference on computer vision}, pages 1348--1357, 2023.

\bibitem[Bucher et~al.(2019)Bucher, Vu, Cord, and P{\'e}rez]{bucher2019zero}
Maxime Bucher, Tuan-Hung Vu, Matthieu Cord, and Patrick P{\'e}rez.
\newblock Zero-shot semantic segmentation.
\newblock \emph{Advances in Neural Information Processing Systems}, 32, 2019.

\bibitem[Caesar et~al.(2018)Caesar, Uijlings, and Ferrari]{caesar2018cvpr}
Holger Caesar, Jasper Uijlings, and Vittorio Ferrari.
\newblock Coco-stuff: Thing and stuff classes in context.
\newblock In \emph{Computer vision and pattern recognition (CVPR), 2018 IEEE conference on}. IEEE, 2018.

\bibitem[Chen et~al.(2016)Chen, Yang, Wang, Xu, and Yuille]{chen2016attention}
Liang-Chieh Chen, Yi Yang, Jiang Wang, Wei Xu, and Alan~L Yuille.
\newblock Attention to scale: Scale-aware semantic image segmentation.
\newblock In \emph{Proceedings of the IEEE conference on computer vision and pattern recognition}, pages 3640--3649, 2016.

\bibitem[Chen et~al.(2017)Chen, Papandreou, Kokkinos, Murphy, and Yuille]{chen2017deeplab}
Liang-Chieh Chen, George Papandreou, Iasonas Kokkinos, Kevin Murphy, and Alan~L Yuille.
\newblock Deeplab: Semantic image segmentation with deep convolutional nets, atrous convolution, and fully connected crfs.
\newblock \emph{IEEE transactions on pattern analysis and machine intelligence}, 40\penalty0 (4):\penalty0 834--848, 2017.

\bibitem[Chen et~al.(2020)Chen, Li, Yu, El~Kholy, Ahmed, Gan, Cheng, and Liu]{chen2020uniter}
Yen-Chun Chen, Linjie Li, Licheng Yu, Ahmed El~Kholy, Faisal Ahmed, Zhe Gan, Yu Cheng, and Jingjing Liu.
\newblock Uniter: Universal image-text representation learning.
\newblock In \emph{European conference on computer vision}, pages 104--120. Springer, 2020.

\bibitem[Cho et~al.(2024)Cho, Shin, Hong, Arnab, Seo, and Kim]{cho2024cat}
Seokju Cho, Heeseong Shin, Sunghwan Hong, Anurag Arnab, Paul~Hongsuck Seo, and Seungryong Kim.
\newblock Cat-seg: Cost aggregation for open-vocabulary semantic segmentation.
\newblock In \emph{Proceedings of the IEEE/CVF Conference on Computer Vision and Pattern Recognition}, pages 4113--4123, 2024.

\bibitem[Ding et~al.(2022{\natexlab{a}})Ding, Xue, Xia, and Dai]{ding2022decoupling}
Jian Ding, Nan Xue, Gui-Song Xia, and Dengxin Dai.
\newblock Decoupling zero-shot semantic segmentation.
\newblock In \emph{Proceedings of the IEEE/CVF Conference on Computer Vision and Pattern Recognition}, pages 11583--11592, 2022{\natexlab{a}}.

\bibitem[Ding et~al.(2022{\natexlab{b}})Ding, Wang, and Tu]{ding2022open}
Zheng Ding, Jieke Wang, and Zhuowen Tu.
\newblock Open-vocabulary universal image segmentation with maskclip.
\newblock \emph{arXiv preprint arXiv:2208.08984}, 2022{\natexlab{b}}.

\bibitem[Esser et~al.(2024)Esser, Kulal, Blattmann, Entezari, M{\"u}ller, Saini, Levi, Lorenz, Sauer, Boesel, et~al.]{esser2024scaling}
Patrick Esser, Sumith Kulal, Andreas Blattmann, Rahim Entezari, Jonas M{\"u}ller, Harry Saini, Yam Levi, Dominik Lorenz, Axel Sauer, Frederic Boesel, et~al.
\newblock Scaling rectified flow transformers for high-resolution image synthesis.
\newblock In \emph{Forty-first international conference on machine learning}, 2024.

\bibitem[Everingham and Winn(2011)]{everingham2011pascal}
Mark Everingham and John Winn.
\newblock The pascal visual object classes challenge 2011 (voc2011) development kit.
\newblock \emph{Pattern Analysis, Statistical Modelling and Computational Learning, Tech. Rep}, 8, 2011.

\bibitem[Ghiasi et~al.(2022)Ghiasi, Gu, Cui, and Lin]{ghiasi2022scaling}
Golnaz Ghiasi, Xiuye Gu, Yin Cui, and Tsung-Yi Lin.
\newblock Scaling open-vocabulary image segmentation with image-level labels.
\newblock In \emph{European Conference on Computer Vision}, pages 540--557. Springer, 2022.

\bibitem[Gu et~al.(2021)Gu, Lin, Kuo, and Cui]{gu2021open}
Xiuye Gu, Tsung-Yi Lin, Weicheng Kuo, and Yin Cui.
\newblock Open-vocabulary object detection via vision and language knowledge distillation.
\newblock \emph{arXiv preprint arXiv:2104.13921}, 2021.

\bibitem[Han et~al.(2023)Han, Liu, Liew, Ding, Liu, Wang, Tang, Yang, Feng, Zhao, et~al.]{han2023global}
Kunyang Han, Yong Liu, Jun~Hao Liew, Henghui Ding, Jiajun Liu, Yitong Wang, Yansong Tang, Yujiu Yang, Jiashi Feng, Yao Zhao, et~al.
\newblock Global knowledge calibration for fast open-vocabulary segmentation.
\newblock In \emph{Proceedings of the IEEE/CVF International Conference on Computer Vision}, pages 797--807, 2023.

\bibitem[He et~al.(2023)He, Jamonnak, Gou, and Ren]{he2023clip}
Wenbin He, Suphanut Jamonnak, Liang Gou, and Liu Ren.
\newblock Clip-s4: Language-guided self-supervised semantic segmentation.
\newblock In \emph{Proceedings of the IEEE/CVF Conference on Computer Vision and Pattern Recognition}, pages 11207--11216, 2023.

\bibitem[Hossain et~al.(2024)Hossain, Siam, Sigal, and Little]{hossain2024visual}
Mir Rayat~Imtiaz Hossain, Mennatullah Siam, Leonid Sigal, and James~J Little.
\newblock Visual prompting for generalized few-shot segmentation: A multi-scale approach.
\newblock In \emph{Proceedings of the IEEE/CVF Conference on Computer Vision and Pattern Recognition}, pages 23470--23480, 2024.

\bibitem[Huynh et~al.(2022)Huynh, Kuen, Lin, Gu, and Elhamifar]{huynh2022open}
Dat Huynh, Jason Kuen, Zhe Lin, Jiuxiang Gu, and Ehsan Elhamifar.
\newblock Open-vocabulary instance segmentation via robust cross-modal pseudo-labeling.
\newblock In \emph{Proceedings of the IEEE/CVF Conference on Computer Vision and Pattern Recognition}, pages 7020--7031, 2022.

\bibitem[Jia et~al.(2021)Jia, Yang, Xia, Chen, Parekh, Pham, Le, Sung, Li, and Duerig]{jia2021scaling}
Chao Jia, Yinfei Yang, Ye Xia, Yi-Ting Chen, Zarana Parekh, Hieu Pham, Quoc Le, Yun-Hsuan Sung, Zhen Li, and Tom Duerig.
\newblock Scaling up visual and vision-language representation learning with noisy text supervision.
\newblock In \emph{International conference on machine learning}, pages 4904--4916. PMLR, 2021.

\bibitem[Jiang et~al.(2023)Jiang, Chen, Zhao, Chen, Ping, Tran, Xu, Zeng, and Chilimbi]{jiang2023understanding}
Qian Jiang, Changyou Chen, Han Zhao, Liqun Chen, Qing Ping, Son~Dinh Tran, Yi Xu, Belinda Zeng, and Trishul Chilimbi.
\newblock Understanding and constructing latent modality structures in multi-modal representation learning.
\newblock In \emph{Proceedings of the IEEE/CVF Conference on Computer Vision and Pattern Recognition}, pages 7661--7671, 2023.

\bibitem[Kirillov et~al.(2019)Kirillov, He, Girshick, Rother, and Doll{\'a}r]{kirillov2019panoptic}
Alexander Kirillov, Kaiming He, Ross Girshick, Carsten Rother, and Piotr Doll{\'a}r.
\newblock Panoptic segmentation.
\newblock In \emph{Proceedings of the IEEE/CVF conference on computer vision and pattern recognition}, pages 9404--9413, 2019.

\bibitem[Krizhevsky et~al.(2012)Krizhevsky, Sutskever, and Hinton]{krizhevsky2012imagenet}
Alex Krizhevsky, Ilya Sutskever, and Geoffrey~E Hinton.
\newblock Imagenet classification with deep convolutional neural networks.
\newblock \emph{Advances in neural information processing systems}, 25, 2012.

\bibitem[Lampert et~al.(2013)Lampert, Nickisch, and Harmeling]{lampert2013attribute}
Christoph~H Lampert, Hannes Nickisch, and Stefan Harmeling.
\newblock Attribute-based classification for zero-shot visual object categorization.
\newblock \emph{IEEE transactions on pattern analysis and machine intelligence}, 36\penalty0 (3):\penalty0 453--465, 2013.

\bibitem[Liang et~al.(2023)Liang, Wu, Dai, Li, Zhao, Zhang, Zhang, Vajda, and Marculescu]{liang2023open}
Feng Liang, Bichen Wu, Xiaoliang Dai, Kunpeng Li, Yinan Zhao, Hang Zhang, Peizhao Zhang, Peter Vajda, and Diana Marculescu.
\newblock Open-vocabulary semantic segmentation with mask-adapted clip.
\newblock In \emph{Proceedings of the IEEE/CVF Conference on Computer Vision and Pattern Recognition}, pages 7061--7070, 2023.

\bibitem[Liang et~al.(2022)Liang, Zhang, Kwon, Yeung, and Zou]{liang2022mind}
Victor~Weixin Liang, Yuhui Zhang, Yongchan Kwon, Serena Yeung, and James~Y Zou.
\newblock Mind the gap: Understanding the modality gap in multi-modal contrastive representation learning.
\newblock \emph{Advances in Neural Information Processing Systems}, 35:\penalty0 17612--17625, 2022.

\bibitem[Lin et~al.(2017)Lin, Doll{\'a}r, Girshick, He, Hariharan, and Belongie]{lin2017feature}
Tsung-Yi Lin, Piotr Doll{\'a}r, Ross Girshick, Kaiming He, Bharath Hariharan, and Serge Belongie.
\newblock Feature pyramid networks for object detection.
\newblock In \emph{Proceedings of the IEEE conference on computer vision and pattern recognition}, pages 2117--2125, 2017.

\bibitem[Liu et~al.(2023)Liu, Bai, Li, Wang, and Tang]{liu2023open}
Yong Liu, Sule Bai, Guanbin Li, Yitong Wang, and Yansong Tang.
\newblock Open-vocabulary segmentation with semantic-assisted calibration.
\newblock \emph{arXiv preprint arXiv:2312.04089}, 2023.

\bibitem[Liu et~al.(2024)Liu, Bai, Li, Wang, and Tang]{liu2024open}
Yong Liu, Sule Bai, Guanbin Li, Yitong Wang, and Yansong Tang.
\newblock Open-vocabulary segmentation with semantic-assisted calibration.
\newblock In \emph{Proceedings of the IEEE/CVF Conference on Computer Vision and Pattern Recognition}, pages 3491--3500, 2024.

\bibitem[Long et~al.(2015)Long, Shelhamer, and Darrell]{long2015fully}
Jonathan Long, Evan Shelhamer, and Trevor Darrell.
\newblock Fully convolutional networks for semantic segmentation.
\newblock In \emph{Proceedings of the IEEE conference on computer vision and pattern recognition}, pages 3431--3440, 2015.

\bibitem[Lu et~al.(2019)Lu, Batra, Parikh, and Lee]{lu2019vilbert}
Jiasen Lu, Dhruv Batra, Devi Parikh, and Stefan Lee.
\newblock Vilbert: Pretraining task-agnostic visiolinguistic representations for vision-and-language tasks.
\newblock \emph{Advances in neural information processing systems}, 32, 2019.

\bibitem[Mokady et~al.(2021)Mokady, Hertz, and Bermano]{mokady2021clipcap}
Ron Mokady, Amir Hertz, and Amit~H Bermano.
\newblock Clipcap: Clip prefix for image captioning.
\newblock \emph{arXiv preprint arXiv:2111.09734}, 2021.

\bibitem[Mottaghi et~al.(2014)Mottaghi, Chen, Liu, Cho, Lee, Fidler, Urtasun, and Yuille]{mottaghi2014role}
Roozbeh Mottaghi, Xianjie Chen, Xiaobai Liu, Nam-Gyu Cho, Seong-Whan Lee, Sanja Fidler, Raquel Urtasun, and Alan Yuille.
\newblock The role of context for object detection and semantic segmentation in the wild.
\newblock In \emph{Proceedings of the IEEE conference on computer vision and pattern recognition}, pages 891--898, 2014.

\bibitem[Mu et~al.(2022)Mu, Kirillov, Wagner, and Xie]{mu2022slip}
Norman Mu, Alexander Kirillov, David Wagner, and Saining Xie.
\newblock Slip: Self-supervision meets language-image pre-training.
\newblock In \emph{European conference on computer vision}, pages 529--544. Springer, 2022.

\bibitem[Mukhoti et~al.(2023)Mukhoti, Lin, Poursaeed, Wang, Shah, Torr, and Lim]{mukhoti2023open}
Jishnu Mukhoti, Tsung-Yu Lin, Omid Poursaeed, Rui Wang, Ashish Shah, Philip~HS Torr, and Ser-Nam Lim.
\newblock Open vocabulary semantic segmentation with patch aligned contrastive learning.
\newblock In \emph{Proceedings of the IEEE/CVF Conference on Computer Vision and Pattern Recognition}, pages 19413--19423, 2023.

\bibitem[Paszke et~al.(2019)Paszke, Gross, Massa, Lerer, Bradbury, Chanan, Killeen, Lin, Gimelshein, Antiga, et~al.]{paszke2019pytorch}
Adam Paszke, Sam Gross, Francisco Massa, Adam Lerer, James Bradbury, Gregory Chanan, Trevor Killeen, Zeming Lin, Natalia Gimelshein, Luca Antiga, et~al.
\newblock Pytorch: An imperative style, high-performance deep learning library.
\newblock \emph{Advances in neural information processing systems}, 32, 2019.

\bibitem[Peng et~al.(2023)Peng, Tian, Wu, Wang, Liu, Su, and Jia]{peng2023hierarchical}
Bohao Peng, Zhuotao Tian, Xiaoyang Wu, Chengyao Wang, Shu Liu, Jingyong Su, and Jiaya Jia.
\newblock Hierarchical dense correlation distillation for few-shot segmentation.
\newblock In \emph{Proceedings of the IEEE/CVF Conference on Computer Vision and Pattern Recognition}, pages 23641--23651, 2023.

\bibitem[Pont-Tuset et~al.(2020)Pont-Tuset, Uijlings, Changpinyo, Soricut, and Ferrari]{pont2020connecting}
Jordi Pont-Tuset, Jasper Uijlings, Soravit Changpinyo, Radu Soricut, and Vittorio Ferrari.
\newblock Connecting vision and language with localized narratives.
\newblock In \emph{Computer Vision--ECCV 2020: 16th European Conference, Glasgow, UK, August 23--28, 2020, Proceedings, Part V 16}, pages 647--664. Springer, 2020.

\bibitem[Radford et~al.(2021)Radford, Kim, Hallacy, Ramesh, Goh, Agarwal, Sastry, Askell, Mishkin, Clark, et~al.]{radford2021learning}
Alec Radford, Jong~Wook Kim, Chris Hallacy, Aditya Ramesh, Gabriel Goh, Sandhini Agarwal, Girish Sastry, Amanda Askell, Pamela Mishkin, Jack Clark, et~al.
\newblock Learning transferable visual models from natural language supervision.
\newblock In \emph{International conference on machine learning}, pages 8748--8763. PMLR, 2021.

\bibitem[Rombach et~al.(2022)Rombach, Blattmann, Lorenz, Esser, and Ommer]{rombach2022high}
Robin Rombach, Andreas Blattmann, Dominik Lorenz, Patrick Esser, and Bj{\"o}rn Ommer.
\newblock High-resolution image synthesis with latent diffusion models.
\newblock In \emph{Proceedings of the IEEE/CVF conference on computer vision and pattern recognition}, pages 10684--10695, 2022.

\bibitem[Schuhmann et~al.(2022)Schuhmann, Beaumont, Vencu, Gordon, Wightman, Cherti, Coombes, Katta, Mullis, Wortsman, et~al.]{schuhmann2022laion}
Christoph Schuhmann, Romain Beaumont, Richard Vencu, Cade Gordon, Ross Wightman, Mehdi Cherti, Theo Coombes, Aarush Katta, Clayton Mullis, Mitchell Wortsman, et~al.
\newblock Laion-5b: An open large-scale dataset for training next generation image-text models.
\newblock \emph{Advances in Neural Information Processing Systems}, 35:\penalty0 25278--25294, 2022.

\bibitem[Shan et~al.(2024)Shan, Wu, Zhu, Shao, Sang, and Gao]{shan2024open}
Xiangheng Shan, Dongyue Wu, Guilin Zhu, Yuanjie Shao, Nong Sang, and Changxin Gao.
\newblock Open-vocabulary semantic segmentation with image embedding balancing.
\newblock In \emph{Proceedings of the IEEE/CVF Conference on Computer Vision and Pattern Recognition}, pages 28412--28421, 2024.

\bibitem[Sharma et~al.(2018)Sharma, Ding, Goodman, and Soricut]{sharma2018conceptual}
Piyush Sharma, Nan Ding, Sebastian Goodman, and Radu Soricut.
\newblock Conceptual captions: A cleaned, hypernymed, image alt-text dataset for automatic image captioning.
\newblock In \emph{Proceedings of the 56th Annual Meeting of the Association for Computational Linguistics (Volume 1: Long Papers)}, pages 2556--2565, 2018.

\bibitem[Simonyan and Zisserman(2014)]{simonyan2014very}
Karen Simonyan and Andrew Zisserman.
\newblock Very deep convolutional networks for large-scale image recognition.
\newblock \emph{arXiv preprint arXiv:1409.1556}, 2014.

\bibitem[Tan and Bansal(2019)]{tan2019lxmert}
Hao Tan and Mohit Bansal.
\newblock Lxmert: Learning cross-modality encoder representations from transformers.
\newblock \emph{arXiv preprint arXiv:1908.07490}, 2019.

\bibitem[Tan and Le(2019)]{tan2019efficientnet}
Mingxing Tan and Quoc Le.
\newblock Efficientnet: Rethinking model scaling for convolutional neural networks.
\newblock In \emph{International conference on machine learning}, pages 6105--6114. PMLR, 2019.

\bibitem[Thomee et~al.(2016)Thomee, Shamma, Friedland, Elizalde, Ni, Poland, Borth, and Li]{thomee2016yfcc100m}
Bart Thomee, David~A Shamma, Gerald Friedland, Benjamin Elizalde, Karl Ni, Douglas Poland, Damian Borth, and Li-Jia Li.
\newblock Yfcc100m: The new data in multimedia research.
\newblock \emph{Communications of the ACM}, 59\penalty0 (2):\penalty0 64--73, 2016.

\bibitem[Wang et~al.(2019)Wang, Liew, Zou, Zhou, and Feng]{wang2019panet}
Kaixin Wang, Jun~Hao Liew, Yingtian Zou, Daquan Zhou, and Jiashi Feng.
\newblock Panet: Few-shot image semantic segmentation with prototype alignment.
\newblock In \emph{proceedings of the IEEE/CVF international conference on computer vision}, pages 9197--9206, 2019.

\bibitem[Wang et~al.(2024{\natexlab{a}})Wang, He, Xuan, Sebastian, Ono, Li, Behpour, Doan, Gou, Shen, and Ren]{10656965}
Xiaoqi Wang, Wenbin He, Xiwei Xuan, Clint Sebastian, Jorge~Piazentin Ono, Xin Li, Sima Behpour, Thang Doan, Liang Gou, Han-Wei Shen, and Liu Ren.
\newblock { USE: Universal Segment Embeddings for Open-Vocabulary Image Segmentation }.
\newblock In \emph{2024 IEEE/CVF Conference on Computer Vision and Pattern Recognition (CVPR)}, pages 4187--4196, Los Alamitos, CA, USA, 2024{\natexlab{a}}. IEEE Computer Society.

\bibitem[Wang et~al.(2024{\natexlab{b}})Wang, He, Xuan, Sebastian, Ono, Li, Behpour, Doan, Gou, Shen, et~al.]{wang2024use}
Xiaoqi Wang, Wenbin He, Xiwei Xuan, Clint Sebastian, Jorge~Piazentin Ono, Xin Li, Sima Behpour, Thang Doan, Liang Gou, Han-Wei Shen, et~al.
\newblock Use: Universal segment embeddings for open-vocabulary image segmentation.
\newblock In \emph{Proceedings of the IEEE/CVF Conference on Computer Vision and Pattern Recognition}, pages 4187--4196, 2024{\natexlab{b}}.

\bibitem[Wang et~al.(2024{\natexlab{c}})Wang, Li, Kallidromitis, Kato, Kozuka, and Darrell]{wang2024hierarchical}
Xudong Wang, Shufan Li, Konstantinos Kallidromitis, Yusuke Kato, Kazuki Kozuka, and Trevor Darrell.
\newblock Hierarchical open-vocabulary universal image segmentation.
\newblock \emph{Advances in Neural Information Processing Systems}, 36, 2024{\natexlab{c}}.

\bibitem[Wu et~al.(2019)Wu, Kirillov, Massa, Lo, and Girshick]{wu2019detectron2}
Yuxin Wu, Alexander Kirillov, Francisco Massa, Wan-Yen Lo, and Ross Girshick.
\newblock Detectron2.
\newblock \url{https://github.com/facebookresearch/detectron2}, 2019.

\bibitem[Xian et~al.(2019)Xian, Choudhury, He, Schiele, and Akata]{xian2019semantic}
Yongqin Xian, Subhabrata Choudhury, Yang He, Bernt Schiele, and Zeynep Akata.
\newblock Semantic projection network for zero-and few-label semantic segmentation.
\newblock In \emph{Proceedings of the IEEE/CVF Conference on Computer Vision and Pattern Recognition}, pages 8256--8265, 2019.

\bibitem[Xie et~al.(2024)Xie, Cao, Xie, Khan, and Pang]{xie2024sed}
Bin Xie, Jiale Cao, Jin Xie, Fahad~Shahbaz Khan, and Yanwei Pang.
\newblock Sed: A simple encoder-decoder for open-vocabulary semantic segmentation.
\newblock In \emph{Proceedings of the IEEE/CVF Conference on Computer Vision and Pattern Recognition}, pages 3426--3436, 2024.

\bibitem[Xu et~al.(2022{\natexlab{a}})Xu, De~Mello, Liu, Byeon, Breuel, Kautz, and Wang]{xu2022groupvit}
Jiarui Xu, Shalini De~Mello, Sifei Liu, Wonmin Byeon, Thomas Breuel, Jan Kautz, and Xiaolong Wang.
\newblock Groupvit: Semantic segmentation emerges from text supervision.
\newblock In \emph{Proceedings of the IEEE/CVF Conference on Computer Vision and Pattern Recognition}, pages 18134--18144, 2022{\natexlab{a}}.

\bibitem[Xu et~al.(2023{\natexlab{a}})Xu, Liu, Vahdat, Byeon, Wang, and De~Mello]{xu2023open}
Jiarui Xu, Sifei Liu, Arash Vahdat, Wonmin Byeon, Xiaolong Wang, and Shalini De~Mello.
\newblock Open-vocabulary panoptic segmentation with text-to-image diffusion models.
\newblock In \emph{Proceedings of the IEEE/CVF Conference on Computer Vision and Pattern Recognition}, pages 2955--2966, 2023{\natexlab{a}}.

\bibitem[Xu et~al.(2022{\natexlab{b}})Xu, Zhang, Wei, Lin, Cao, Hu, and Bai]{xu2022simple}
Mengde Xu, Zheng Zhang, Fangyun Wei, Yutong Lin, Yue Cao, Han Hu, and Xiang Bai.
\newblock A simple baseline for open-vocabulary semantic segmentation with pre-trained vision-language model.
\newblock In \emph{European Conference on Computer Vision}, pages 736--753. Springer, 2022{\natexlab{b}}.

\bibitem[Xu et~al.(2023{\natexlab{b}})Xu, Zhang, Wei, Hu, and Bai]{xu2023side}
Mengde Xu, Zheng Zhang, Fangyun Wei, Han Hu, and Xiang Bai.
\newblock Side adapter network for open-vocabulary semantic segmentation.
\newblock In \emph{Proceedings of the IEEE/CVF Conference on Computer Vision and Pattern Recognition}, pages 2945--2954, 2023{\natexlab{b}}.

\bibitem[Yang et~al.(2023)Yang, Huang, Wei, Peng, Jiang, Jiang, Wei, Wang, Hu, Qiu, et~al.]{yang2023attentive}
Yifan Yang, Weiquan Huang, Yixuan Wei, Houwen Peng, Xinyang Jiang, Huiqiang Jiang, Fangyun Wei, Yin Wang, Han Hu, Lili Qiu, et~al.
\newblock Attentive mask clip.
\newblock In \emph{Proceedings of the IEEE/CVF International Conference on Computer Vision}, pages 2771--2781, 2023.

\bibitem[Yao et~al.(2021)Yao, Huang, Hou, Lu, Niu, Xu, Liang, Li, Jiang, and Xu]{yao2021filip}
Lewei Yao, Runhui Huang, Lu Hou, Guansong Lu, Minzhe Niu, Hang Xu, Xiaodan Liang, Zhenguo Li, Xin Jiang, and Chunjing Xu.
\newblock Filip: Fine-grained interactive language-image pre-training.
\newblock \emph{arXiv preprint arXiv:2111.07783}, 2021.

\bibitem[Yu et~al.(2023)Yu, He, Deng, Shen, and Chen]{yu2023convolutions}
Qihang Yu, Ju He, Xueqing Deng, Xiaohui Shen, and Liang-Chieh Chen.
\newblock Convolutions die hard: Open-vocabulary segmentation with single frozen convolutional clip.
\newblock \emph{Advances in Neural Information Processing Systems}, 36:\penalty0 32215--32234, 2023.

\bibitem[Zhang et~al.(2024)Zhang, Li, Guo, and Wang]{zhang2024sair}
Canyu Zhang, Xiaoguang Li, Qing Guo, and Song Wang.
\newblock Sair: Learning semantic-aware implicit representation.
\newblock In \emph{European Conference on Computer Vision}, pages 319--335. Springer, 2024.

\bibitem[Zhang et~al.(2021)Zhang, Kang, Yang, and Wei]{zhang2021few}
Gengwei Zhang, Guoliang Kang, Yi Yang, and Yunchao Wei.
\newblock Few-shot segmentation via cycle-consistent transformer.
\newblock \emph{Advances in Neural Information Processing Systems}, 34:\penalty0 21984--21996, 2021.

\bibitem[Zhang and Agrawala(2024)]{zhang2024transparentimagelayerdiffusion}
Lvmin Zhang and Maneesh Agrawala.
\newblock Transparent image layer diffusion using latent transparency, 2024.

\bibitem[Zhang and Saligrama(2016)]{zhang2016zero}
Ziming Zhang and Venkatesh Saligrama.
\newblock Zero-shot learning via joint latent similarity embedding.
\newblock In \emph{proceedings of the IEEE Conference on Computer Vision and Pattern Recognition}, pages 6034--6042, 2016.

\bibitem[Zhao et~al.(2022)Zhao, Wu, Wu, Zhang, and Wang]{zhao2022crossmodal}
Ziyu Zhao, Zhenyao Wu, Xinyi Wu, Canyu Zhang, and Song Wang.
\newblock Crossmodal few-shot 3d point cloud semantic segmentation.
\newblock In \emph{Proceedings of the 30th ACM international conference on multimedia}, pages 4760--4768, 2022.

\bibitem[Zhao et~al.(2024)Zhao, Cai, Zhang, Li, and Wang]{zhao2024crossmodal}
Ziyu Zhao, Pingping Cai, Canyu Zhang, Xiaoguang Li, and Song Wang.
\newblock Crossmodal few-shot 3d point cloud semantic segmentation via view synthesis.
\newblock In \emph{Proceedings of the 32nd ACM International Conference on Multimedia}, pages 8777--8785, 2024.

\bibitem[Zhong et~al.(2022)Zhong, Yang, Zhang, Li, Codella, Li, Zhou, Dai, Yuan, Li, et~al.]{zhong2022regionclip}
Yiwu Zhong, Jianwei Yang, Pengchuan Zhang, Chunyuan Li, Noel Codella, Liunian~Harold Li, Luowei Zhou, Xiyang Dai, Lu Yuan, Yin Li, et~al.
\newblock Regionclip: Region-based language-image pretraining.
\newblock In \emph{Proceedings of the IEEE/CVF conference on computer vision and pattern recognition}, pages 16793--16803, 2022.

\bibitem[Zhou et~al.(2017)Zhou, Zhao, Puig, Fidler, Barriuso, and Torralba]{zhou2017scene}
Bolei Zhou, Hang Zhao, Xavier Puig, Sanja Fidler, Adela Barriuso, and Antonio Torralba.
\newblock Scene parsing through ade20k dataset.
\newblock In \emph{Proceedings of the IEEE conference on computer vision and pattern recognition}, pages 633--641, 2017.

\bibitem[Zhou et~al.(2022)Zhou, Loy, and Dai]{zhou2022extract}
Chong Zhou, Chen~Change Loy, and Bo Dai.
\newblock Extract free dense labels from clip.
\newblock In \emph{European Conference on Computer Vision}, pages 696--712. Springer, 2022.

\bibitem[Zou et~al.(2023)Zou, Dou, Yang, Gan, Li, Li, Dai, Behl, Wang, Yuan, et~al.]{zou2023generalized}
Xueyan Zou, Zi-Yi Dou, Jianwei Yang, Zhe Gan, Linjie Li, Chunyuan Li, Xiyang Dai, Harkirat Behl, Jianfeng Wang, Lu Yuan, et~al.
\newblock Generalized decoding for pixel, image, and language.
\newblock In \emph{Proceedings of the IEEE/CVF Conference on Computer Vision and Pattern Recognition}, pages 15116--15127, 2023.

\end{thebibliography}
